\documentclass[runningheads]{llncs}
\usepackage{graphicx}
\usepackage{comment}
\usepackage{amsmath,amssymb} 
\usepackage{color}
\usepackage[pagebackref=true,breaklinks=true,letterpaper=true,colorlinks,bookmarks=false]{hyperref}

\usepackage{booktabs}
\usepackage{multirow}
\usepackage{makecell}
\usepackage{fixltx2e}
\usepackage{tabularx}
\usepackage{array}
\usepackage{caption} 
\usepackage{enumitem}
\usepackage{wrapfig}

\let\llncssubparagraph\subparagraph
\let\subparagraph\paragraph
\usepackage[compact]{titlesec}
\let\subparagraph\llncssubparagraph

\newcommand{\eg}{\emph{e.g.}}

\begin{document}
\pagestyle{headings}
\mainmatter
\def\ECCVSubNumber{2668}  

\title{Attentive Normalization} 

\titlerunning{Attentive Normalization}
%
\author{Xilai Li \and
Wei Sun \and
Tianfu Wu~\thanks{T. Wu is the corresponding author.}}
\authorrunning{Li, Sun and Wu}
%
\institute{Department of Electrical and Computer Engineering, NC State University \\
\email{\{xli47, wsun12, tianfu\_wu\}@ncsu.edu}}
\maketitle

\begin{abstract}
In state-of-the-art deep neural networks, both feature normalization  and feature attention  have become ubiquitous. 
They are usually studied as separate modules, however. In this paper, we propose a light-weight integration between the two schema and present Attentive Normalization (AN). Instead of learning a single affine transformation, AN learns a mixture of affine transformations and utilizes their weighted-sum as the final affine transformation applied to re-calibrate features in an instance-specific way. The weights are learned by leveraging channel-wise feature attention. In experiments, we test the proposed AN using four representative neural architectures 
in the ImageNet-1000 classification benchmark and the MS-COCO 2017 object detection and instance segmentation benchmark. AN obtains consistent performance improvement for different neural architectures in both benchmarks with absolute increase of top-1 accuracy in ImageNet-1000 between 0.5\% and 2.7\%, and absolute increase up to 1.8\% and 2.2\% for  bounding box and mask AP  in MS-COCO respectively. We observe that the proposed AN provides a strong alternative to the widely used Squeeze-and-Excitation (SE) module. The source codes are publicly available at \href{https://github.com/iVMCL/AOGNet-v2}{the ImageNet Classification Repo} and \href{https://github.com/iVMCL/AttentiveNorm\_Detection}{the MS-COCO Detection and Segmentation Repo}.
\end{abstract}


\section{Introduction} 
Pioneered by Batch Normalization (BN)~\cite{BatchNorm}, feature normalization has become ubiquitous in the development of deep learning. Feature normalization consists of two components: \textit{feature standardization} and \textit{channel-wise affine transformation}. The latter is introduced to provide the capability of undoing the standardization (by design), and can be treated as \textit{feature re-calibration} in general. Many variants of BN have been proposed for practical deployment in terms of variations of training and testing settings with remarkable progress obtained. They can be roughly divided into two categories: 
\begin{figure*} [t]
    \centering
    \includegraphics[width=0.9\linewidth]{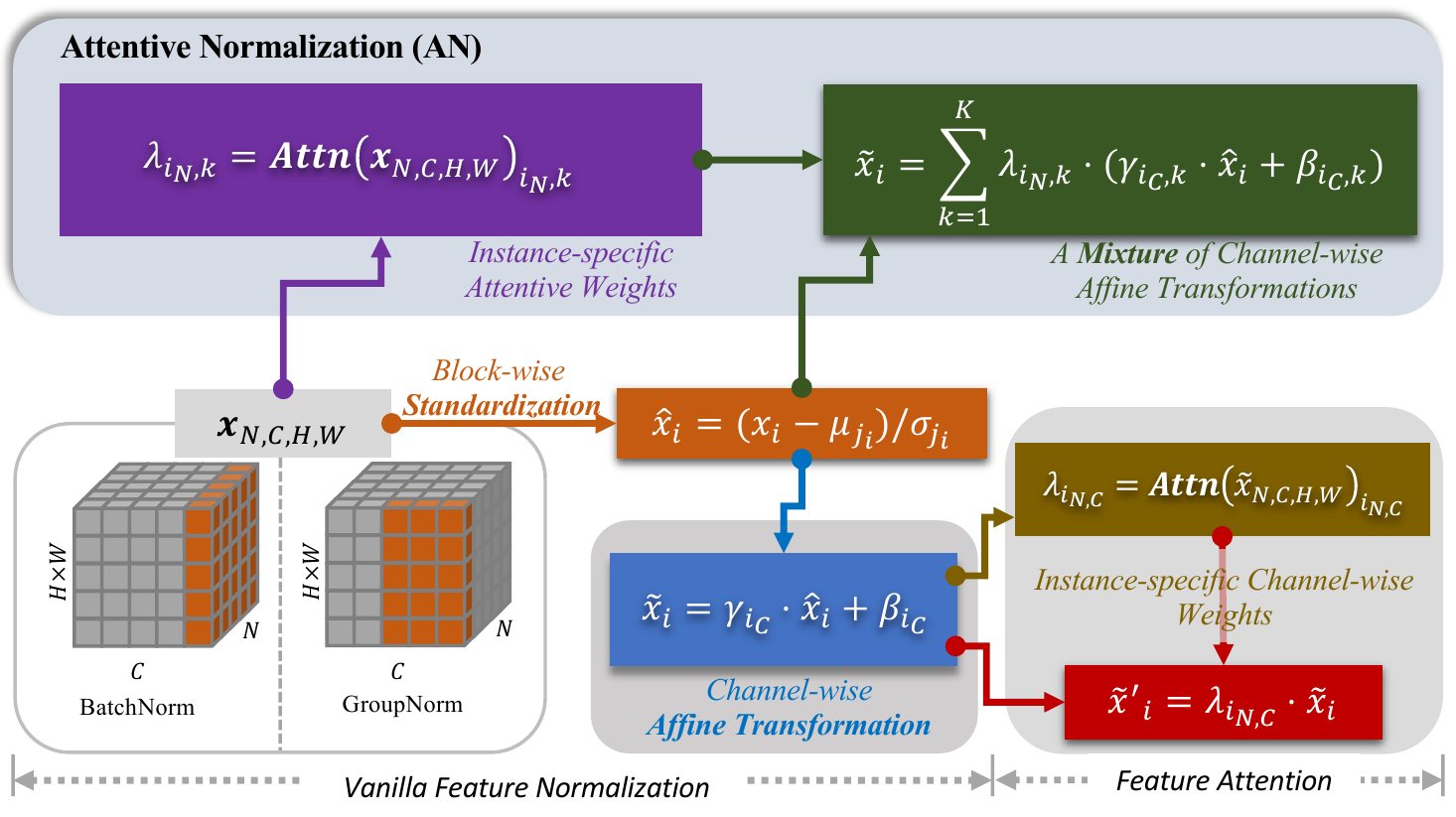}
    \caption{Illustration of the proposed Attentive Normalization (AN). AN aims to harness the best of a base feature normalization (e.g., BN or GN) and channel-wise feature attention in a single light-weight module.     See text for details.}
    \label{fig:AN} 
\end{figure*}

\textit{i) Generalizing feature standardization.} Different methods are proposed for computing the mean and standard deviation or for modeling/whitening the data distribution in general, within a min-batch. They include Batch Renormalization~\cite{BatchReNorm}, Decorrelated BN~\cite{DecorBN},  Layer Normalization (LN)~\cite{LayerNorm}, Instance Normalization (IN)~\cite{InstNorm}, Instance-level Meta Normalization~\cite{InstMetaNorm},  Group Normalization (GN)~\cite{GroupNorm},  Mixture Normalization~\cite{MixtureNorm} and Mode Normalization~\cite{ModeNorm}. Switchable Normalization (SN)~\cite{SwitchNorm} and its sparse variant (SSN)~\cite{SSN} learn to switch between different vanilla schema. These methods adopt the vanilla channel-wise affine transformation after standardization, and are often proposed for discriminative learning tasks. 

\textit{ii) Generalizing feature re-calibration.} Instead of treating the affine transformation parameters directly as model parameters, different types of task-induced conditions (\eg, class labels in conditional image synthesis using generative adversarial networks) are leveraged and encoded as latent vectors, which are then used to learn the affine transformation parameters, including different conditional BNs~\cite{CBatchNorm1,CBatchNorm2,ConditionNorm,CGAN,BigGAN}, style-adaptive IN~\cite{StyleGAN} or layout-adaptive IN~\cite{SPatialAdaNorm,ISLANorm}. These methods have been mainly proposed in generative learning tasks, except for the recently proposed Instance-level Meta Normalization~\cite{InstMetaNorm} in discriminative learning tasks. 

In the meanwhile, \textit{feature attention} has also become an indispensable mechanism for improving task performance in deep learning. For computer vision, spatial attention is inherently captured by convolution operations within short-range context, and by non-local extensions~\cite{NonlocalNet,CrissCross} for long-range context. Channel-wise attention is relatively less exploited. The squeeze-and-excitation (SE) unit~\cite{SENet} is one of the most popular designs, which learn instance-specific channel-wise attention weights to re-calibrate an input feature map. Unlike the affine transformation parameters in feature normalization, the attention weights for re-calibrating an feature map are often directly learned from the input feature map in the spirit of self-attention, and often instance-specific or pixel-specific.

Although both feature normalization  and feature attention  have become ubiquitous in state-of-the-art DNNs, they are usually studied as separate modules. Therefore, in this paper we address the following problem: 
\textit{How to learn to re-calibrate feature maps in a way of harnessing the best of feature normalization and feature attention in a single light-weight module?} 
And, we present \textbf{Attentive Normalization (AN)}: Fig.~\ref{fig:AN} illustrates the proposed AN. The basic idea is straightforward. Conceptually, the affine transformation component in feature normalization (Section~\ref{sec:featnorm}) and the re-scaling computation in feature attention  play the same role in learning-to-re-calibrate an input feature map, thus providing the foundation for integration (Section~\ref{sec:featattn}). More specifically, consider a feature normalization backbone such as BN or GN, our proposed AN keeps the block-wise standardization component unchanged. Unlike the vanilla feature normalization in which the affine transformation parameters ($\gamma$'s and $\beta$'s) are often frozen in testing, we want the affine transformation parameters to be adaptive and dynamic in both training and testing,  controlled directly by the input feature map. The intuition behind doing so is that it will be more flexible in accounting for different statistical discrepancies between training and testing in general, and between different sub-populations caused by underlying inter-/intra-class variations in the data.   

To achieve the dynamic and adaptive control of affine transformation parameters, the proposed AN utilizes a simple design (Section~\ref{sec:AN}). It learns a mixture of $K$ affine transformations and exploits feature attention mechanism to learn the instance-specific weights for the $K$ components. The final affine transformation used to re-calibrate an input feature map is the weighted sum of the learned $K$ affine transformations. We propose a general formulation for the proposed AN and study how to learn the weights in an efficient and effective way (Section~\ref{sec:learn_wts}).

\section{Related Work}
\textbf{Feature Normalization.}
There are two types of normalization schema, feature normalization (including raw data)~\cite{BatchNorm,BatchReNorm,LayerNorm,InstNorm,GroupNorm,SwitchNorm,SSN,MixtureNorm,ModeNorm} and weight normalization~\cite{WN,OrthWtsNorm}. Unlike the former, the latter is to normalize model parameters to decouple the magnitudes of parameter vectors from their directions.
We focus on feature normalization in this paper. 

Different feature normalization schema differ in how the mean and variance are computed. BN~\cite{BatchNorm} computes the channel-wise mean and variance in the entire min-batch which is driven by improving training efficiency and model generalizability. BN has been deeply analyzed in terms of how it helps optimization~\cite{HowBNWorks}. DecorBN~\cite{DecorBN} utilizes a whitening operation (ZCA) to go beyond the centering and scaling in the vanilla BN. BatchReNorm~\cite{BatchReNorm} introduces extra parameters to control the pooled mean and variance to reduce BN's dependency on the batch size. IN~\cite{InstNorm} focuses on channel-wise and instance-specific statistics which stems from the task of artistic image style transfer. LN~\cite{LayerNorm} computes the instance-specific mean and variance from all channels which is designed to help optimization in recurrent neural networks (RNNs). GN~\cite{GroupNorm} stands in the sweet spot between LN and IN focusing on instance-specific and channel-group-wise statistics, especially when only small batches are applicable in practice. In practice, synchronized BN~\cite{SyncBN} across multiple GPUs becomes increasingly favorable against GN in some applications. SN~\cite{SwitchNorm} leaves the design choices of feature normalization schema to the learning system itself by computing weighted sum integration of BN, LN, IN and/or GN via softmax, showing more flexible applicability, followed by SSN~\cite{SSN} which learns to make exclusive selection. Instead of computing one mode (mean and variance), MixtureNorm~\cite{MixtureNorm} introduces a mixture of Gaussian densities to approximate the data distribution in a mini-batch. ModeNorm~\cite{ModeNorm} utilizes a general form of multiple-mode computation. Unlike those methods, the proposed AN focuses on generalizing the affine transformation component. Related to our work, Instance-level Meta normalization(ILM)~\cite{InstMetaNorm} first utilizes an encoder-decoder sub-network to learn affine transformation parameters and then add them together to the model's affine transformation parameters. Unlike ILM, the proposed AN utilizes a mixture of affine transformations and leverages feature attention to learn the instance-specific attention weights.

On the other hand, conditional feature normalization schema~\cite{CBatchNorm1,CBatchNorm2,ConditionNorm,BigGAN,StyleGAN,SPatialAdaNorm} \cite{ISLANorm} have been developed and shown remarkable progress in conditional and unconditional image synthesis. Conditional BN learns condition-specific affine transformations in terms of conditions such as class labels,  image style, label maps and geometric layouts. Unlike those methods, the proposed AN learns self-attention data-driven weights for mixture components of affine transformations.

\textbf{Feature Attention.}
Similar to feature normalization, feature attention is also an important building block in the development of deep learning. Residual Attention Network~\cite{ResAttention} uses a trunk-and-mask joint spatial and channel attention module in an encoder-decoder style for improving performance. To reduce the computational cost, channel and spatial attention are separately applied in~\cite{CBAM}. The SE module~\cite{SENet} further simplifies the attention mechanism by developing a light-weight  channel-wise attention method. The proposed AN leverages the idea of SE in learning attention weights, but formulates the idea in a novel way.

\textbf{Our Contributions.} This paper makes three main contributions: (i) It presents Attentive Normalization which harnesses the best of feature normalization and feature attention (channel-wise). To our knowledge, AN is the first work that studies self-attention based conditional and adaptive feature normalization in visual recognition tasks.  
(ii) It presents a lightweight integration method for deploying AN in different widely used building blocks of ResNets, DenseNets, MobileNetsV2 and AOGNets. 
(iii) It obtains consistently better results than the vanilla feature normalization backbones by a large margin across different neural architectures in two large-scale benchmarks, ImageNet-1000 and MS-COCO.

\section{The Proposed Attentive Normalization}\label{sec:AN} 
In this section, we present details of the proposed attentive normalization. 
Consider a DNN for 2D images, denote by $\mathbf{x}$ a feature map with axes in the convention order of $(N, C, H, W)$ (i.e., batch, channel, height and width). $\mathbf{x}$ is represented by a 4D tensor. Let $i=(i_N, i_C, i_H, i_W)$ be the address index in the 4D tensor. $\mathbf{x}_i$ represents the feature response at a position $i$.

\subsection{Background on Feature Normalization}\label{sec:featnorm}
Existing feature normalization schema often consist of two components (Fig.~\ref{fig:AN}):  

\textit{i) Block-wise Standardization}.  Denote by $B_j$ a block (slice) in a given 4-D tensor $\mathbf{x}$. For example, for BN, we have $j = 1, \cdots, C$ and $B_j=\{\mathbf{x}_i | \forall i, i_C=j\}$. We first compute the empirical mean and standard deviation in $B_j$, denoted by $\mu_j$ and $\sigma_j$ respectively: $\mu_j = \frac{1}{M}\sum_{x\in B_j} x,\quad 
        \sigma_j = \sqrt{\frac{1}{M}\sum_{x\in B_j} (x-\mu_j)^2 + \epsilon}$,  
where $M=|B_j|$ and $\epsilon$ is a small positive constant to ensure $\sigma_j > 0$ for the sake of numeric stability. Then, let $j_i$ be the index of the block that the position $i$ belongs to, and we standardize the feature response by, 
\begin{equation}
    \hat{\mathbf{x}}_i = \frac{1}{\sigma_{j_i}} (\mathbf{x}_i - \mu_{j_i})
\end{equation}

\textit{ii) Channel-wise Affine Transformation}. Denote by $\gamma_c$ and $\beta_c$ the scalar coefficient (re-scaling) and offset (re-shifting)  parameter respectively for the $c$-th channel.  The re-calibrated feature response at a position $i$ is then computed by, 
\begin{equation}
    \Tilde{\mathbf{x}}_i = \gamma_{i_C} \cdot \hat{\mathbf{x}}_i + \beta_{i_C}, \label{eq:featnorm}
\end{equation}
where $\gamma_c$'s and $\beta_c$'s are shared by all the instances in a min-batch across the spatial domain. They are usually frozen in testing and fine-tuning. 

\subsection{Background on Feature Attention}\label{sec:featattn}
We focus on channel-wise attention and briefly review the Squeeze-Excitation (SE) module~\cite{SENet}. SE usually takes the feature normalization result (Eqn.~\ref{eq:featnorm}) as its input (the bottom-right of Fig.~\ref{fig:AN}), and learns channel-wise attention weights: 

\textit{i) The squeeze module} encodes the inter-dependencies between feature channels  in a low dimensional latent space with the reduction rate $r$ (e.g., $r=16$), 
    \begin{equation}
        S(\Tilde{\mathbf{x}};\theta_S) = v,\, v\in \mathbb{R}^{N\times \frac{C}{r}\times 1\times 1}, \label{eq:squeeze}
    \end{equation}
    which is implemented by a sub-network consisting of a global average pooling layer (AvgPool), a fully-connected (FC) layer and rectified linear unit (ReLU)~\cite{AlexNet}. $\theta_S$ collects all the model parameters. 

\textit{ii) The excitation module} computes the channel-wise attention weights, denoted by $\lambda$, by decoding the learned latent representations $v$, 
\begin{equation}
    E(v;\theta_E)=\lambda,\, \lambda\in \mathbb{R}^{N\times C\times 1 \times 1},
\end{equation}
which is implemented by a sub-network consisting of a FC layer and a sigmoid layer. $\theta_E$ collects all model parameters.

Then, the input, $\Tilde{\mathbf{x}}$ is re-calibrated by, 
\begin{align}
    \Tilde{\mathbf{x}}^{SE}_i = \lambda_{i_N, i_C} \cdot \Tilde{\mathbf{x}}_i = (\lambda_{i_N, i_C} \cdot \gamma_{i_C}) \cdot \hat{\mathbf{x}}_i + \lambda_{i_N, i_C} \cdot \beta_{i_C}, \label{eq:se} 
\end{align}    
where the second step is obtained by plugging in Eqn.~\ref{eq:featnorm}. \textbf{It is thus straightforward to see the foundation facilitating the integration between feature normalization and channel-wise feature attention.}
However, the SE module often entails a significant number of extra parameters (e.g., $\sim$2.5M extra parameters for ResNet50~\cite{ResidualNet} which originally consists of $\sim$25M parameters, resulting in 10\% increase). We aim to design more parsimonious integration that can further improve performance.
    
\subsection{Attentive Normalization}\label{sec:attnnorm}
Our goal is to generalize Eqn.~\ref{eq:featnorm} in re-calibrating feature responses to enable dynamic and adaptive control in both training and testing. On the other hand, our goal is to simplify Eqn.~\ref{eq:se} into a single light-weight module, rather than, for example, the two-module setup using BN+SE. In general, we have,
\begin{equation}
     \Tilde{\mathbf{x}}^{AN}_i = \Gamma(\mathbf{x};\theta_{\Gamma})_i \cdot \hat{\mathbf{x}}_i + \mathbb{B}(\mathbf{x};\theta_{\mathbb{B}})_i, \label{eq:recalibration}
\end{equation}
where both $\Gamma(\mathbf{x};\theta_{\Gamma})$ and $\mathbb{B}(\mathbf{x};\theta_{\mathbb{B}})$ are functions of the entire input feature map (without standardization~\footnote{We tried the variant of learning $\Gamma()$ and $\mathbb{B}()$ from the standardized features and observed it works worse, so we ignore it in our experiments.}) with parameters $\theta_{\Gamma}$ and $\theta_{\mathbb{B}}$ respectively. They both compute 4D tensors of the size same as the input feature map and can be parameterized by some attention guided light-weight DNNs. The subscript in $\Gamma(\mathbf{x};\theta_{\Gamma})_i$ and $\mathbb{B}(\mathbf{x};\theta_{\mathbb{B}})_i$ represents the learned re-calibration weights at a position $i$. 

In this paper, we focus on learning instance-specific channel-wise affine transformations. To that end, we have three components as follows. 

\textit{i) Learning a Mixture of $K$ Channel-wise Affine Transformations.} Denote by $\gamma_{k,c}$ and $\beta_{k,c}$ the re-scaling and re-shifting (scalar) parameters respectively for the $c$-th channel in the $k$-th mixture component. They are model parameters learned end-to-end via back-propagation. 

\textit{ii) Learning Attention Weights for the $K$ Mixture Components.}  Denote by $\lambda_{n, k}$ the instance-specific mixture component weight ($n\in [1, N]$ and $k\in [1, K]$), and by $\lambda$ the $N\times K$ weight matrix. $\lambda$ is learned via some attention-guided function from the entire input feature map, 
\begin{equation}
    \lambda = A(\mathbf{x}; \theta_{\lambda}), \label{eq:attn_wts}
\end{equation}
where $\theta_{\lambda}$ collects all the parameters. 

\textit{iii) Computing the Final Affine Transformation.} With the learned $\gamma_{k,c}$, $\beta_{k,c}$ and $\lambda$, the re-calibrated feature response is computed by, 
\begin{equation}
     \Tilde{\mathbf{x}}^{AN}_i = \sum_{k=1}^K \lambda_{i_N, k} [\gamma_{k, i_C}\cdot \hat{\mathbf{x}}_i + \beta_{k, i_C}], \label{eq:recalibration1}
\end{equation}
where $\lambda_{i_N, k}$ is shared by the re-scaling parameter and the re-shifting parameter for simplicity. Since the attention weights $\lambda$ are adaptive and dynamic in both training and testing, the proposed AN realizes adaptive and dynamic feature re-calibration. Compared to the general form (Eqn.~\ref{eq:recalibration}), we have,
\begin{equation}
    \Gamma(\mathbf{x})_i= \sum_{k=1}^K \lambda_{i_N, k}\cdot \gamma_{k,i_C},\,\,  
    \mathbb{B}(\mathbf{x})_i=\sum_{k=1}^K \lambda_{i_N, k}\cdot \beta_{k,i_C}.
\end{equation} 

Based on the formulation, there are \textbf{a few advantages of the proposed AN in training, fine-tuning and testing }a DNN: 
\begin{itemize}
\item The channel-wise affine transformation parameters, $\gamma_{k,i_C}$'s and $\beta_{k,i_C}$'s, are shared across spatial dimensions and by data instances, which can learn population-level knowledge in a more fine-grained manner than a single affine transformation in the vanilla feature normalization. 
\item $\lambda_{i_N, k}$'s are instance specific and learned from features that are not standardized. Combining them with $\gamma_{k,i_C}$'s and $\beta_{k,i_C}$'s (Eqn.~\ref{eq:recalibration1}) enables AN paying attention to both the population (what the common and useful information are) and the individuals (what the specific yet critical information are). The latter is particularly useful for testing samples slightly ``drifted" from training population, that is to improve generalizability. 
Their  weighted sum encodes more direct and ``actionable" information for re-calibrating standardized features (Eqn.~\ref{eq:recalibration1}) without being delayed until back-propagation updates as done in the vanilla feature normalization. 
\item  In fine-tuning, especially between different tasks (\eg, from image classification to object detection), $\gamma_{k,i_C}$'s and $\beta_{k,i_C}$'s are usually frozen as done in the vanilla feature normalization. They carry information from a source task. But, $\theta_{\lambda}$ (Eqn.~\ref{eq:attn_wts}) are allowed to be fine-tuned, thus potentially better realizing transfer learning for a target task. 
This is a desirable property since we can decouple training correlation between tasks. For example, when GN~\cite{GroupNorm} is applied in object detection in MS-COCO, it is fine-tuned from a feature backbone with GN trained in ImageNet, instead of the one with BN that usually has better performance in ImageNet. 
As we shall show in experiments, the proposed AN facilitates a smoother transition. We can use the proposed AN (with BN) as the normalization backbone in pre-training in ImageNet, and then use AN (with GN) as the normalization backbone for the head classifiers in MS-COCO with significant improvement.    
\end{itemize}

\subsubsection{Details of Learning Attention Weights}\label{sec:learn_wts} 
We present a simple method for computing the attention weights $A(\mathbf{x}; \theta_{\lambda})$ (Eqn.~\ref{eq:attn_wts}). Our goal is to learn a weight coefficient for each component from each individual instance in a mini-batch (i.e, a $N\times K$ matrix). 
The question of interest is how to characterize the underlying importance of a channel $c$ from its realization across the spatial dimensions $(H, W)$ in an instance, such that we will learn a more informative instance-specific weight coefficient for a channel $c$ in re-calibrating the feature map $\mathbf{x}$.

In realizing Eqn.~\ref{eq:attn_wts}, the proposed method is similar in spirit to the squeeze module in SENets~\cite{SENet} to maintain light-weight implementation. To show the difference, let's first rewrite the vanilla squeeze module (Eqn.~\ref{eq:squeeze}), 
 \begin{equation}
        v=S(\mathbf{x};\theta_S) = ReLU(fc(AvgPool(\mathbf{x});\theta_S))\, ,  \label{eq:squeeze1}
    \end{equation}
where the mean of a channel $c$ (via global average pooling, $AvgPool(\cdot)$) is used to characterize its underlying importance. We generalize this assumption by taking into account both mean and standard deviation empirically computed for a channel $c$, denoted by $\mu_c$ and $\sigma_c$ respectively. More specifically, we compare three different designs using: 
\begin{enumerate}
    \item [i)] The mean $\mu_c$ only as done in SENets.  
    \item [ii)] The concatenation of the mean and standard deviation, $(\mu_c, \sigma_c)$. 
    \item [iii)] The coefficient of variation or the relative standard deviation (RSD), $\frac{\sigma_c}{\mu_c}$. RSD measures the dispersion of an underlying distribution (i.e., the extent to which the distribution is stretched or squeezed) which intuitively conveys more information in learning attention weights for re-calibration.
\end{enumerate}
 RSD is indeed observed to work better in our experiments\footnote{In implementation, we use the reverse $\frac{\mu_c}{\sigma_c + \epsilon}$ for numeric stability, which is equivalent to the original formulation when combing with the $fc$ layer.}. Eqn.~\ref{eq:attn_wts} is then expanded  with two choices,   
\begin{align}
   \textit{Choice 1: } A_1(\mathbf{x}; \theta_{\lambda}) &= Act(fc(RSD(\mathbf{x});\theta_{\lambda})), \label{eq:attn_wts1} \\
   \nonumber \textit{Choice 2: } A_2(\mathbf{x}; \theta_{\lambda}) &= Act(BN(fc(RSD(\mathbf{x});\theta_{fc});\theta_{BN})), 
\end{align}
where $Act(\cdot)$ represents a non-linear activation function for which we compare three designs: 
\begin{itemize}
\item [i)] The vanilla $ReLU(\cdot)$ as used in the squeeze module of SENets.
\item [ii)] The vanilla $sigmoid(\cdot)$ as used in the excitation module of SENets.  
\item [iii)] The channel-wise $softmax(\cdot)$. 
\item [iv)] The piece-wise linear hard analog of the sigmoid function, so-called $hsigmoid$ function~\cite{mobilenetv3}, $hsigmoid(a) = \min(\max(a+3.0, 0), 6.0) / 6.0$.
\end{itemize}
The $hsigmoid(\cdot)$ is observed to work better in our experiments. In the Choice 2 (Eqn.~\ref{eq:attn_wts1}), we apply the vanilla BN~\cite{BatchNorm} after the FC layer, which normalizes the learned attention weights across all the instances in a mini-batch with the hope of balancing the instance-specific attention weights better. The Choice 2 improves performance in our experiments in ImageNet.

In AN, we have another hyper-parameter, $K$. For stage-wise building block based neural architectures such as the four neural architectures tested in our experiments, we use different $K$'s for different stages with smaller values for early stages. For example, for the 4-stage setting, we typically use $K=10, 10, 20, 20$ for the four stages respectively based on our ablation study. The underlying assumption is that early stages often learn low-to-middle level features which are considered to be shared more between different categories, while later stages learn more category-specific features which may entail larger mixtures.

\section{Experiments}\label{sec:exp}
In this section, we first show the ablation study verifying the design choices in the proposed AN. Then, we present detailed comparisons and analyses.

\begin{figure} [t]
    \centering
    \includegraphics[width=0.9\linewidth]{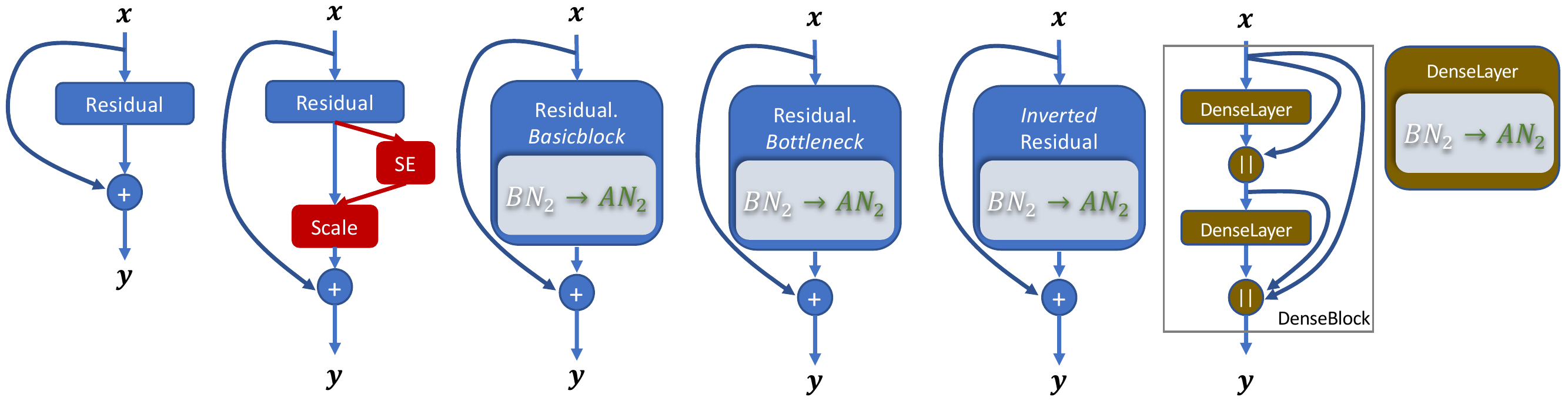}
    \caption{Illustration of integrating the proposed AN in different building blocks. The first two show the vanilla residual block and the SE-residual block. The remaining four are: the Basicblock and Bottleneck design of a residual block, the inverted residual block (used in MobileNetV2), and the DenseBlock. 
    For the residual block and its variants, the proposed AN is used to replace the vanilla BN(s) followed the last $3\times 3$ convolution in different blocks. This potentially enables  jointly  integrating  local  spatial  attention (conveyed by the $3\times 3$ convolution) in  learning the instance-specific attention weights, which is also observed helpful in~\cite{SWhitenning} and is shown beneficial for the SE module itself in our experiments (Table~\ref{table:se-results}). For the dense block, we replace the second vanilla BN (after the $1\times 1$ convolution applied to the concatenated features) with our AN. 
    }
    \label{fig:ResBlockAN}
\end{figure}

\textbf{Data and Evaluation Metric.} We use two benchmarks, the ImageNet-1000 classification benchmark (ILSVRC2012)~\cite{ImageNet} and the MS-COCO object detection and instance segmentation benchmark~\cite{COCO}. The ImageNet-1000 benchmark consists of about $1.28$ million images for training, and $50,000$ for validation, from $1,000$ classes. We apply a single-crop with size $224\times224$ in evaluation. Following the common protocol, we report  the top-1 and top-5 classification error rates tested using a single model on the validation set. For the MS-COCO benchmark, there are 80 categories of objects. We use {\tt train2017} in training and evaluate the trained models using {\tt val2107}. We report the standard COCO metrics of Average Precision (AP) at different intersection-over-union (IoU) thresholds, \eg, AP$_{50}$ and AP$_{75}$, for bounding box detection (AP$^{bb}_{IoU}$) and instance segmentation (AP$^m_{IoU}$), and the mean AP over IoU=$0.5:0.05:0.75$, AP$^{bb}$ and AP$^m$ for bounding box detection and instance segmentation respectively.


\textbf{Neural Architectures and Vanilla Feature Normalization Backbones.} We use four representative neural architectures: (i) \textit{ResNets}~\cite{ResidualNet} (ResNet50 and ResNet101), which are the most widely used architectures in practice, (ii) \textit{DenseNets}~\cite{DenseNet}, which are popular alternatives to ResNets, (iii) \textit{MobileNetV2}~\cite{sandler2018mobilenetv2}. MobileNets are popular architectures under mobile settings and MobileNetV2 uses inverted residuals and linear Bottlenecks, and (iv) \textit{AOGNets}~\cite{AOGNets}, which are grammar-guided networks and represent an interesting direction of network architecture engineering with better performance than ResNets and DenseNets. So, the improvement by our AN will be both broadly useful for existing ResNets, DenseNets and MobileNets based deployment in practice and potentially insightful for on-going and future development of more advanced and more powerful DNNs in the community. 

In classification, we use BN~\cite{BatchNorm} as the feature normalization backbone for our proposed AN, denoted by \textbf{AN (w/ BN)}. We compare with the vanilla BN, GN~\cite{GroupNorm} and SN~\cite{SwitchNorm}. In object detection and instance segmentation, we use the Mask-RCNN framework~\cite{maskrcnn} and its cascade variant~\cite{cascadercnn} in the MMDetection code platform~\cite{mmdetection}. We fine-tune feature backbones pretrained on the ImageNet-1000 dataset. We also test the proposed AN using GN as the feature normalization backbone, denoted by \textbf{AN (w/ GN)} in the head classifier of Mask-RCNN.

\textbf{Where to Apply AN?} Fig.~\ref{fig:ResBlockAN} illustrates the integration of our proposed AN in different building blocks. At the first thought, it is straightforward to replace all vanilla feature normalization modules (\eg, BN) in a DNN. It may not be necessary to do so, similar in spirit to the SE-residual block which re-calibrates the residual part once in a building block. As we shall see, our ablation study supports the design choice shown in Fig.~\ref{fig:ResBlockAN}.     

\textbf{Initialization of our AN.} The initialization of $\gamma_{k,c}$'s and $\beta_{k,c}$'s (Eqn.~\ref{eq:recalibration1}) is based on, $\gamma_{k, c} = 1.0 + \mathcal{N}(0, 1) \times 0.1$ and $\beta_{k,c}=\mathcal{N}(0, 1)\times 0.1$, where $\mathcal{N}(0, 1)$ represents the standard Gaussian distribution. This type of initialization is also adopted for conditional BN used in the BigGAN~\cite{BigGAN}.

\subsection{Ablation Study}\label{sec:ablation} \begin{wraptable}{r}{0.6\textwidth} 
    \centering
    \small{
    \resizebox{0.6\textwidth}{!}{
    \begin{tabular}{r|cccc}
    \hline 
    Design Choices in AN (w/ BN) & \#Params & FLOPS & top-1 & top-5 \\ \hline
\textbf{mean} + $A_2(\cdot)$ + hsigmoid + $K=\bigl(\begin{smallmatrix} {10}\\ {10}\\ {20}\\{20}
\end{smallmatrix}\bigr)$ & 25.76M & 4.09G & 21.85 & 5.92 \\ \hline 
 \textbf{(mean,std)} + $A_2(\cdot)$ + hsigmoid + $K=\bigl(\begin{smallmatrix} {10}\\ {10}\\ {20}\\{20}
\end{smallmatrix}\bigr)$ & 25.82M & 4.09G & 21.73 & 5.85 \\ \hline
 RSD + ${\textbf{A}_1}(\cdot)$ + hsigmoid + $K=\bigl(\begin{smallmatrix} {10}\\ {10}\\ {20}\\{20}
\end{smallmatrix}\bigr)$ & 25.76M & 4.09G & 21.76 & 6.05 \\ \hline 
 RSD + $A_2(\cdot)$ + \textbf{softmax} + $K=\bigl(\begin{smallmatrix} {10}\\ {10}\\ {20}\\{20}
\end{smallmatrix}\bigr)$ & 25.76M & 4.09G & 21.72 & 5.90 \\ \hline
 RSD + $A_2(\cdot)$ + \textbf{relu} + $K=\bigl(\begin{smallmatrix} {10}\\ {10}\\ {20}\\{20} \end{smallmatrix}\bigr)$ & 25.96M & 4.09G & 21.89 & 6.04 \\ \hline
 RSD + $A_2(\cdot)$ + \textbf{sigmoid} + $K=\bigl(\begin{smallmatrix} {10}\\ {10}\\ {20}\\{20}
\end{smallmatrix}\bigr)$ & 25.76M & 4.09G & 21.96 & 5.91 \\ \hline
 RSD + $A_2(\cdot)$ + hsigmoid + $\textbf{K}=\bigl(\begin{smallmatrix} {5}\\ {5}\\ {10}\\{10}
\end{smallmatrix}\bigr)$ & 25.76M & 4.09G & 21.92 & 5.93 \\ \hline
 RSD + $A_2(\cdot)$ + hsigmoid + $\textbf{K}=\bigl(\begin{smallmatrix} {20}\\ {20}\\ {40}\\{40} \end{smallmatrix}\bigr)$ & 25.96M & 4.09G & 21.62 & 5.63 \\ \hline
  RSD + $A_2(\cdot)$ + hsigmoid + $K=\bigl(\begin{smallmatrix} {10}\\ {10}\\ {20}\\{20}
\end{smallmatrix}\bigr)$ & 25.76M & 4.09G & \textbf{21.59} & \textbf{5.58} \\ \bottomrule
 \textbf{*} RSD + $A_2(\cdot)$ + hsigmoid + $K=\bigl(\begin{smallmatrix} {10}\\ {10}\\ {20}\\{20}
\end{smallmatrix}\bigr)$ & 26.96M & 4.10G & 22.15 & 6.24 \\ \bottomrule
    \end{tabular} }}
    \\ [1ex]
    \caption{Ablation study on different design choices in AN with BN as feature normalization backbone using ResNet50+Bottleneck in ImageNet-1000. 
    * means AN is applied to all the BNs of the network. }\label{table:ablation} 
\end{wraptable} 
We compare different design choices in our proposed AN using ResNet50 in  ImageNet-1000. Table~\ref{table:ablation} summarizes the results. There are four categories of design choices: The first three are related to the realization of learning the attention weights (Eqn.~\ref{eq:attn_wts}): three types of inputs, two architectural choices and four activation function choices. The last one refers to the number $K$ of components in the mixture of affine transformation which is used for each of the four stages in ResNet50 and we empirically select three options for simplicity.  All the models are trained using the same settings (the vanilla setup in Section~\ref{sec:results_imagenet}). 

\textbf{The best combination} is RSD + $A_2(\cdot)$ + hsigmoid + $K=\bigl(\begin{smallmatrix} {10}\\ {10}\\ {20}\\{20}
\end{smallmatrix}\bigr)$. During our development, we first observed the best combination based on our intuitive reasoning and small experiments (a few epochs) in the process, and then design this ablation study to verify the design choices. Based on the observed best combination, we further verify that \textit{replacing all vanilla BNs is not helpful} (the last row in Table~\ref{table:ablation}). One explanation is that we may not need to re-calibrate the features using our AN (as well as other channel-wise feature attention methods) for both before and after a $1\times 1$ convolution, since channel-wise re-calibration can be tackled by the $1\times 1$ convolution kernel and the vanilla feature normalization themselves in training.  
The ablation study is in support of the intuitions and design choices discussed in Section~\ref{sec:learn_wts}. 

\subsection{Image Classification in ImageNet-1000}\label{sec:results_imagenet}
\textit{Common Training Settings.}
We use 8 GPUs (NVIDIA V100) to train models \begin{wraptable}{r}{0.5\textwidth} 
    \centering
    \small{
    \resizebox{0.49\textwidth}{!}{
    \begin{tabular}{lrccc}
    \toprule 
    \multicolumn{5}{c}{\textit{The Vanilla Setup}} \\ \hline
    Method & \#Params & FLOPS & top-1 & top-5 \\ \hline
    ResNet34+BN & 21.80M & 3.68G & 25.58$_{\downarrow (1.15)}$ & 8.19$_{\downarrow (0.76)}$ \\
    \textbf{ResNet34+AN} & 21.92M & 3.68G & \textbf{24.43} & \textbf{7.43}   \\\bottomrule
    ResNet50-BN  & 25.56M & 4.09G & 23.01$_{\downarrow  (1.42)}$ & 6.68$_{\downarrow  (0.80)}$  \\ 
    $^\dagger$ResNet50-GN~\cite{GroupNorm} & 25.56M & 4.09G & 23.52$_{\downarrow  (1.93)}$ & 6.85$_{\downarrow  (0.97)}$  \\
    $^\dagger$ResNet50-SN~\cite{SwitchNorm} & 25.56M & - & 22.43$_{\downarrow  (0.83)}$ & 6.35$_{\downarrow  (0.47)}$ \\
    $^\dagger$ResNet50-SE~\cite{SENet} & 28.09M & 4.12G & 22.37$_{\downarrow  (0.78)}$ & 6.36$_{\downarrow  (0.48)}$ \\
    ResNet50-SE & 28.09M & 4.12G & 22.35$_{\downarrow  (0.76)}$ & 6.09$_{\downarrow  (0.21)}$ \\
    \textbf{ResNet50-AN} & 25.76M & 4.09G & \textbf{21.59} & \textbf{5.88}  \\
    \bottomrule
    ResNet101-BN & 44.57M & 8.12G & 21.33$_{\downarrow (0.72)}$ & 5.85$_{\downarrow (0.44)}$  \\
    \textbf{ResNet101-AN} & 45.00M & 8.12G & \textbf{20.61} & \textbf{5.41}  \\\bottomrule
    DenseNet121-BN & 7.98M & 2.86G & 25.35$_{\downarrow (2.73)}$ & 7.83$_{\downarrow (1.41)}$ \\ 
    \textbf{DenseNet121-AN} & 8.34M & 2.86G & \textbf{22.62} & \textbf{6.42}   \\ \bottomrule
    MobileNetV2-BN & 3.50M & 0.34G & 28.69$_{\downarrow (2.02)}$ & 9.33$_{\downarrow (0.77)}$  \\
    \textbf{MobileNetV2-AN} & 3.56M & 0.34G & \textbf{26.67} & \textbf{8.56}  \\ \bottomrule
    AOGNet12M-BN & 12.26M & 2.19G  & 22.22$_{\downarrow (0.94)}$ & 6.06$_{\downarrow (0.30)}$  \\ 
    \textbf{AOGNet12M-AN} & 12.37M & 2.19G & \textbf{21.28} & \textbf{5.76}  \\ \hline
    AOGNet40M-BN & 40.15M & 7.51G & 19.84$_{\downarrow (0.51)}$ & 4.94$_{\downarrow (0.22)}$ \\
    \textbf{AOGNet40M-AN} & 40.39M & 7.51G & \textbf{19.33} & \textbf{4.72}   \\ \bottomrule \bottomrule
    
    \multicolumn{5}{c}{\textit{The State-of-the-Art Setup}} \\ \hline
    Method & \#Params & FLOPS & top-1 & top-5 \\ \hline
    ResNet50-BN & 25.56M & 4.09G & 21.08$_{\downarrow (1.16)}$ & 5.56$_{\downarrow (0.52)}$  \\ 
    \textbf{ResNet50-AN} & 25.76M & 4.09G & \textbf{19.92} & \textbf{5.04}  \\    \bottomrule
    ResNet101-BN & 44.57M & 8.12G & 19.71$_{\downarrow (0.86)}$ & 4.89$_{\downarrow (0.26)}$  \\
    \textbf{ResNet101-AN} & 45.00M & 8.12G & \textbf{18.85} & \textbf{4.63}  \\\bottomrule
    AOGNet12M-BN & 12.26M & 2.19G & 21.63$_{\downarrow (1.06)}$ & 5.60$_{\downarrow (0.22)}$ \\
    \textbf{AOGNet12M-AN} & 12.37M & 2.19G & \textbf{20.57} & \textbf{5.38} \\ \bottomrule
    AOGNet40M-BN & 40.15M & 7.51G & 18.70$_{\downarrow (0.57)}$ & 4.47$_{\downarrow (0.21)}$ \\
    \textbf{AOGNet40M-AN} & 40.39M & 7.51G & \textbf{18.13} & \textbf{4.26} \\ \bottomrule
    \end{tabular}  }}
    \caption{Comparisons between BN and our AN (w/ BN) in terms of the top-1 and top-5 error rates (\%) in the ImageNet-1000 validation set using \textit{the vanilla setup} and \textit{the state-of-the-art setup}. 
    $^\dagger$ means the model is not trained by us. 
    All other models are trained from scratch under the same settings. 
    }\label{table:imagenet-results} 
\end{wraptable}  using the same settings for apple-to-apple comparisons. The method proposed in~\cite{KaimingNormInit} is used to initialize all convolutions for all models. The batch size is 128 per GPU.
with FP16 optimization used in training to reduce the training time.
The mean and standard deviation for block-wise standardization are computed \textit{within} each GPU. The initial learning rate is $0.4$, and the cosine learning rate scheduler~\cite{cosine_lr} is used with $5$ warm-up epochs and weight decay $1\times10^{-4}$ and momentum $0.9$. For AN, the best practice observed in our ablation study (Table~\ref{table:ablation}) is used. AN is not used in the stem layer in all the models.   In addition to the common settings, we have two different setups in experimental comparisons:

\textit{i) The Vanilla Setup.} We adopt the basic data augmentation scheme (random crop and horizontal flip) in training as done in~\cite{ResidualNet}. We train the models for 120 epochs. All ResNets~\cite{ResidualNet} use the vanilla stem layer with $7\times 7$ convolution. The MobileNetsV2 uses $3\times 3$ convolution in the stem layer. The AOGNets use two consecutive $3\times 3$ convolution in the stem layer. All the $\gamma$ and $\beta$ parameters of the feature normalization backbones are initialized to $1$ and $0$ respectively.    

\textit{ii) The State-of-the-Art Setup.} There are different aspects in the vanilla setup which have better variants developed with better performance shown~\cite{BagofTricksImgCls}. \textit{We want to address whether the improvement by our proposed AN are truly fundamental or will disappear with more advanced tips and tricks added in training ConvNets.} First, on top of the basic data augmentation, we also use label smoothing~\cite{LabelSmoothing} (with rate 0.1) and the mixup (with rate 0.2)~\cite{mixup}. We increase the total number of epochs to 200. We use the same stem layer with two consecutive $3\times 3$ convolution for all models. For ResNets, we add the zero $\gamma$ initialization trick, which uses 0 to initialize the last normalization layer  to make the initial state of a residual block to be identity. 

\textbf{Results Summary.} Table~\ref{table:imagenet-results} shows the comparison results for the two setups respectively. \textbf{Our proposed AN consistently obtains the best top-1 and top-5 accuracy results with more than 0.5\% absolute top-1 accuracy increase (up to 2.7\%) in all models without bells and whistles.} \textit{The improvement is often obtained with negligible extra parameters} (e.g., 0.06M parameter increase in MobileNetV2 for  2.02\% absolute top-1 accuracy increase, and 0.2M parameter increase in ResNet50 with 1.42\% absolute top-1 accuracy increase) \textit{at almost no extra computational cost} (up to the precision used in measuring FLOPs). With ResNet50, our AN also outperforms
GN~\cite{GroupNorm} and SN~\cite{SwitchNorm} by 1.93\% and 0.83\% in top-1 accuracy respectively. For GN, it is known that it works (slightly) worse than BN under the normal (big) mini-batch setting~\cite{GroupNorm}. For SN, \begin{wraptable}{r}{0.5\textwidth} 
    \centering
    \small{
    \resizebox{0.49\textwidth}{!}{
    \begin{tabular}{lrccc}
    \hline 
    Method & \#Params & FLOPS & top-1 & top-5 \\ \hline
    ResNet50-SE (BN$_3$) & 28.09M & 4.12G & 22.35$_{\downarrow  (0.76)}$ & 6.09$_{\downarrow  (0.21)}$ \\
    ResNet50-SE (BN$_2$) & 26.19M & 4.12G & 22.10$_{\downarrow  (0.55)}$ & 6.02$_{\downarrow  (0.14)}$ \\
    ResNet50-SE (All) & 29.33M & 4.13G & 22.13$_{\downarrow (0.52)}$ & $5.96_{\downarrow (0.08)}$ \\ \hline
    ResNet50-AN (w/BN$_3$) & 26.35M & 4.11G & 21.78$_{\downarrow (0.19)}$ & 5.98$_{\downarrow (0.1)}$ \\
    \textbf{ResNet50-AN} (w/BN$_2$) & \textbf{25.76M} & \textbf{4.09G} & \textbf{21.59} & \textbf{5.88}  \\
    ResNet50-AN (All) & 25.92M & 4.10G & 21.85$_{\downarrow (0.26)}$ & 6.06$_{\downarrow (0.18)}$ \\
    \bottomrule
\end{tabular}  }}
\caption{Comparisons between SE and our AN (w/ BN) in terms of the top-1 and top-5 error rates (\%) in the ImageNet-1000 validation set using \textit{the vanilla setup}. By ``(All)", it means SE or AN is used for all the three BNs in a bottleneck block.}\label{table:se-results} 
\end{wraptable} our result shows that it is more beneficial to improve the re-calibration component than to learn-to-switch between different feature normalization schema. 
We observe that the proposed AN is more effective for small ConvNets in terms of performance gain. Intuitively, this makes sense. Small ConvNets usually learn less expressive features. With the mixture of affine transformations and the instance-specific channel-wise feature re-calibration, the proposed AN offers the flexibility of clustering intra-class data better while separating inter-class data  better in training.

\textbf{Comparisons with the SE module.} Our proposed AN provides a strong alternative to the widely used SE module. Table~\ref{table:se-results} shows the comparisons. We observe that applying SE after the second BN in the bottleneck in ResNet50 is also beneficial with better performance and smaller number of extra parameters. 

\begin{table*} [t] 
    \centering
    \small{
    \resizebox{0.9\textwidth}{!}{
    \begin{tabular}{l|ll|c|ccc|ccc}
    Architecture & Backbone & Head & \#Params & AP$^{bb}$ & AP$^{bb}_{50}$ & AP$^{bb}_{75}$ & AP$^{m}$ & AP$^{m}_{50}$ & AP$^{m}_{75}$ \\ \toprule
    \multirow{2}{*}{\rotatebox[origin=c]{0}{MobileNetV2}} & $\mathbb{BN}$ & - & 22.72M & 34.2$_{\downarrow {(1.8)}}$ & 54.6$_{\downarrow {(2.4)}}$ & 37.1$_{\downarrow {(1.8)}}$ & 30.9$_{\downarrow {(1.6)}}$ & 51.1$_{\downarrow {(2.7)}}$ & 32.6$_{\downarrow {(1.9)}}$ \\
    & AN (w/ BN) & -  & 22.78M & \textbf{36.0} & \textbf{57.0} & \textbf{38.9} & \textbf{32.5} & \textbf{53.8} & \textbf{34.5} \\  \bottomrule

    \multirow{5}{*}{\rotatebox[origin=c]{0}{ResNet50}} 
    & $\mathbb{BN}$ & - & 45.71M & 39.2$_{\downarrow {(1.6)}}$ & 60.0$_{\downarrow {(2.1)}}$ & 43.1$_{\downarrow {(1.4)}}$ & 35.2$_{\downarrow {(1.2)}}$ & 56.7$_{\downarrow {(2.2)}}$ & 37.6$_{\downarrow {(1.1)}}$\\ 
    &  $\mathbb{BN}+SE(BN_3)$ & - & 48.23M & 40.1$_{\downarrow {(0.7)}}$ & 61.2$_{\downarrow {(0.9)}}$ & 43.8$_{\downarrow {(0.7)}}$ & 35.9$_{\downarrow {(0.5)}}$ & 57.9$_{\downarrow {(1.0)}}$ & 38.1$_{\downarrow {(0.6)}}$ \\
        &  $\mathbb{BN}+SE(BN_2)$ & - & 46.34M & 40.1$_{\downarrow {(0.7)}}$ & 61.2$_{\downarrow {(0.9)}}$ & 43.8$_{\downarrow {(0.7)}}$ & 35.9$_{\downarrow {(0.5)}}$ & 57.9$_{\downarrow {(1.0)}}$ & 38.4$_{\downarrow {(0.3)}}$ \\
    & AN (w/ BN) & - & 45.91M    & \textbf{40.8} & \textbf{62.1} & \textbf{44.5} & \textbf{36.4} & \textbf{58.9} & \textbf{38.7}  \\ \cline{2-10}
    & $^\dagger$GN  & GN~\cite{GroupNorm} & 45.72M & 40.3$_{\downarrow {(1.3)}}$ & 61.0$_{\downarrow {(1.0)}}$ & 44.0$_{\downarrow {(1.7)}}$ & 35.7$_{\downarrow {(1.7)}}$ & 57.9$_{\downarrow {(1.6)}}$ & 37.7$_{\downarrow {(2.2)}}$ \\
    & $^\dagger$SN & SN~\cite{SwitchNorm} & - & 41.0$_{\downarrow {(0.6)}}$ & \textbf{62.3}$_{\downarrow {(-0.3)}}$ & 45.1$_{\downarrow {(0.6)}}$ & 36.5$_{\downarrow {(0.9)}}$ & 58.9$_{\downarrow {(0.6)}}$ & 38.7$_{\downarrow {(1.2)}}$ \\
    & AN (w/ BN) & AN (w/ GN) & 45.96M & \textbf{41.6} & 62.0 & \textbf{45.7} & \textbf{37.4} & \textbf{59.5} & \textbf{39.9} \\ \bottomrule
    \multirow{4}{*}{\rotatebox[origin=c]{0}{ResNet101}} 
    & $\mathbb{BN}$  & - & 64.70M & 41.4$_{\downarrow {(1.7)}}$ & 62.0$_{\downarrow {(2.1)}}$ & 45.5$_{\downarrow {(1.8)}}$ & 36.8$_{\downarrow {(1.4)}}$ & 59.0$_{\downarrow {(2.0)}}$ & 39.1$_{\downarrow {(1.6)}}$ \\ 
    & AN (w/ BN) & - & 65.15M     & \textbf{43.1} & \textbf{64.1} & \textbf{47.3} & \textbf{{38.2}} & \textbf{61.0} & \textbf{40.7 } \\ \cline{2-10} 
    & $^\dagger$GN  & GN~\cite{GroupNorm} & 64.71M & 41.8$_{\downarrow {(1.4)}}$ & 62.5$_{\downarrow {(1.5)}}$ & 45.4$_{\downarrow {(1.9)}}$ & 36.8$_{\downarrow {(2.0)}}$ & 59.2$_{\downarrow {(2.1)}}$ & 39.0$_{\downarrow {(2.6)}}$ \\
    & AN (w/ BN) & AN (w/ GN) & 65.20M & \textbf{43.2} & \textbf{64.0} & \textbf{47.3} & \textbf{38.8} & \textbf{61.3} & \textbf{41.6} \\ \bottomrule
    \multirow{3}{*}{\rotatebox[origin=c]{0}{AOGNet12M}} & $\mathbb{BN}$ & - & 33.09M & 40.7$_{\downarrow {(1.3)}}$  & 61.4$_{\downarrow {(1.7)}}$ & 44.6$_{\downarrow {(1.5)}}$ & 36.4$_{\downarrow {(1.4)}}$ & 58.4$_{\downarrow {(1.7)}}$ & 38.8$_{\downarrow {(1.6)}}$  \\ 
    & AN (w/ BN) & - & 33.21M & {42.0}$_{\downarrow {(1.0)}}$& 63.1$_{\downarrow {(1.1)}}$ & 46.1$_{\downarrow {(0.7)}}$ & {37.8}$_{\downarrow {(0.9)}}$ & 60.1$_{\downarrow {(1.0)}}$ & 40.4$_{\downarrow {(1.3)}}$ \\ 
    & AN (w/ BN) & AN (w/ GN) & 33.26M & \textbf{43.0} & \textbf{64.2} & \textbf{46.8} & \textbf{38.7} & \textbf{61.1} & \textbf{41.7} \\ \bottomrule
    \multirow{3}{*}{\rotatebox[origin=c]{0}{AOGNet40M}} & $\mathbb{BN}$ & - & 60.73M & 43.4$_{\downarrow {(0.7)}}$ & 64.2$_{\downarrow {(0.9)}}$ & 47.5$_{\downarrow {(0.7)}}$ & 38.5$_{\downarrow {(0.5)}}$ & 61.0$_{\downarrow {(1.0)}}$ & 41.4$_{\downarrow {(0.4)}}$  \\ 
    & AN (w/ BN) & - & 60.97M & {44.1}$_{\downarrow {(0.8)}}$ & 65.1$_{\downarrow {(1.1)}}$ & 48.2$_{\downarrow {(0.9)}}$ & {39.0}$_{\downarrow {(1.2)}}$ & 62.0$_{\downarrow {(1.2)}}$ & 41.8$_{\downarrow {(1.5)}}$ \\
    & AN (w/ BN) & AN (w/ GN) & 61.02M & \textbf{44.9} & \textbf{66.2} & \textbf{49.1} & \textbf{40.2} & \textbf{63.2} & \textbf{43.3} \\ \bottomrule
    \end{tabular} 
    }    
    }
    \caption{Detection and segmentation results in MS-COCO {\tt val2017}~\cite{COCO}. All models use 2x lr scheduling (180k iterations).  $\mathbb{BN}$ means BN is frozen in fine-tuning for object detection. $^\dagger$ means that models are not trained by us. All other models are trained from scratch under the same settings. The numbers show sequential improvement in the two AOGNet models indicating the importance of adding our AN in the backbone and the head respectively. 
    }\label{table:coco-results} 
\end{table*}

\subsection{Object Detection and Segmentation in COCO}
In object detection and segmentation, high-resolution input images are beneficial and often entailed for detecting medium to small objects, but limit the batch-size in training (often 1 or 2 images per GPU). GN~\cite{GroupNorm} and SN~\cite{SwitchNorm} have shown significant progress in handling the applicability discrepancies of feature normalization schema from ImageNet to MS-COCO.    
{We test our AN in MS-COCO following the standard protocol, as done in GN~\cite{GroupNorm}. We build on the MMDetection code platform~\cite{mmdetection}. We observe further performance improvement.} 

We first summarize the details of implementation. Following the terminologies used in MMDetection~\cite{mmdetection}, there are four modular components in the R-CNN detection framework~\cite{FastRCNN,FasterRCNN,maskrcnn}: 
\textit{i) Feature Backbones}. We use the pre-trained networks in Table~\ref{table:imagenet-results} (with the vanilla setup) for fair comparisons in detection, since we compare with some models which are not trained by us from scratch and use the feature backbones pre-trained in a way similar to our vanilla setup and with on par top-1 accuracy. In fine-tuning a network with AN (w/ BN) pre-trained in ImageNet such as ResNet50+AN (w/ BN) in Table~\ref{table:imagenet-results}, we freeze the stem layer and the first stage as commonly done in practice. For the remaining stages, we freeze the standardization component only (the learned mixture of affine transformations and the learned running mean and standard deviation), but allow the attention weight sub-network to be fine-tuned. 
\textit{ii) Neck Backbones}: We test the feature pyramid network (FPN)~\cite{FPN} which is widely used in practice. 
\textit{iii) Head Classifiers}. We test two setups: 
   \textit{ (a) The vanilla setup} as done in GN~\cite{GroupNorm} and SN~\cite{SwitchNorm}. In this setup, we further have two settings: with vs without feature normalization in the bounding box head classifier. The former is denoted by ``-" in Table~\ref{table:coco-results}, and the latter is denoted by the corresponding type of feature normalization scheme in Table~\ref{table:coco-results} (\eg, GN, SN and AN (w/ GN)). We experiment on using AN (w/ GN) in the bounding box head classifier and keeping GN in the mask head unchanged for simplicity. Adding AN (w/ GN) in the mask head classifier may further help improve the performance. When adding AN (w/ GN) in the bounding box head, we adopt the same design choices except for ``Choice 1, $A_1(\cdot)$" (Eqn.~\ref{eq:attn_wts1}) used in learning attention weights.  
    \textit{(b) The state-of-the-art setup } which is based on the cascade generalization of head classifiers~\cite{cascadercnn} and does not include feature normalization scheme, also denoted by ``-" in Table~\ref{table:coco-csc-results}.  
\textit{iv) RoI Operations}. We test the RoIAlign operation~\cite{maskrcnn}. 

 \begin{table*} [t]
    \centering
    \small{
    \resizebox{0.9\textwidth}{!}{
    \begin{tabular}{l|ll|r|ccc|ccc}
     Architecture & Backbone & Head & \#Params & AP$^{bb}$ & AP$^{bb}_{50}$ & AP$^{bb}_{75}$ & AP$^{m}$ & AP$^{m}_{50}$ & AP$^{m}_{75}$ \\ \hline
     \multirow{2}{*}{ResNet101} 
     & $\mathbb{BN}$ & - & 96.32M & 44.4$_{\downarrow (1.4)}$ & 62.5$_{\downarrow (1.8)}$ & 48.4$_{\downarrow (1.4)}$ & 38.2$_{\downarrow (1.4)}$ & 59.7$_{\downarrow (2.0)}$ & 41.3$_{\downarrow (1.4)}$ \\
     & AN (w/ BN)& - & 96.77M & \textbf{45.8} & \textbf{64.3} & \textbf{49.8} & \textbf{39.6} & \textbf{61.7} & \textbf{42.7} \\  \hline
     \multirow{2}{*}{AOGNet40M} & $\mathbb{BN}$& - & 92.35M & 45.6$_{\downarrow (0.9)}$ & 63.9$_{\downarrow (1.1)}$ & 49.7$_{\downarrow (1.1)}$ & 39.3$_{\downarrow (0.7)}$ & 61.2$_{\downarrow (1.1)}$ & 42.7$_{\downarrow (0.4)}$ \\
     & AN (w/ BN)& - & 92.58M & \textbf{46.5} & \textbf{65.0} & \textbf{50.8} & \textbf{40.0} & \textbf{62.3} & \textbf{43.1} \\  \hline
    \end{tabular} }}
    \caption{Results in MS-COCO using the cascade variant~\cite{cascadercnn} of Mask R-CNN. 
    }\label{table:coco-csc-results} 
\end{table*}

\textbf{Result Summary.} The results are summarized in Table~\ref{table:coco-results} and Table~\ref{table:coco-csc-results}. Compared with the vanilla BN that are frozen in fine-tuning, our AN (w/ BN) improves performance by a large margin in terms of both bounding box AP and mask AP (\textit{1.8\% $\&$ 1.6\%} for MobileNetV2, \textit{1.6\% $\&$ 1.2\%} for ResNet50, \textit{1.7\% $\&$ 1.4\%} for ResNet101, \textit{1.3\% $\&$ 1.4\%} for AOGNet12M and \textit{0.7\% $\&$ 0.5\%} for AOGNet40M). It shows the advantages of the self-attention based dynamic and adaptive control of the mixture of affine transformations (although they themselves are frozen) in fine-tuning. 

 With the AN further integrated in the bounding box head classifier of Mask-RCNN and trained from scratch, we also obtain better performance than GN and SN. Compared with the vanilla GN~\cite{GroupNorm}, our AN (w/ GN) improves bounding box and mask AP by 1.3\% and 1.7\% for ResNet50, and 1.4\% and 2.2\% for ResNet101. Compared with SN~\cite{SwitchNorm} which outperforms the vanilla GN in ResNet50, our AN (w/ GN) is also better by 0.6\% bounding box AP and 0.9\%  mask AP increase respectively. Slightly less improvements are observed with AOGNets.   
Similar in spirit to the ImageNet experiments, we want to verify whether the advantages of our AN will disappear if we use state-of-the-art designs for head classifiers of R-CNN such as the widely used cascade R-CNN~\cite{cascadercnn}. Table~\ref{table:coco-csc-results} shows that similar improvements are obtained with ResNet101 and AOGNet40M.

\section{Conclusion}
This paper presents Attentive Normalization (AN) that aims to harness the best of feature normalization and feature attention in a single lightweight module. AN learns a mixture of affine transformations and uses the weighted sum via a self-attention module for re-calibrating standardized features in a dynamic and adaptive way. AN provides a strong alternative to the Squeeze-and-Excitation (SE) module. In experiments, AN is tested with BN and GN as the feature normalization backbones. AN is tested in both ImageNet-1000  and   MS-COCO using four representative networks (ResNets, DenseNets, MobileNetsV2 and AOGNets). It consistently obtains better performance, often by a large margin, than the vanilla feature normalization schema and some state-of-the-art variants.

\section*{Acknowledgement} 
This work is supported in part by NSF IIS-1909644, ARO Grant W911NF1810295, NSF IIS-1822477 and NSF IUSE-2013451.
The views presented in this paper are those of the authors and should not be interpreted as representing any funding agencies.

%
%
\bibliographystyle{splncs04}
\bibliography{egbib}
\end{document}


\pagestyle{headings}
\mainmatter
\def\ECCVSubNumber{2668}  

\title{Supplementary Material for Attentive Normalization}

\titlerunning{ECCV-20 submission ID \ECCVSubNumber} 
\authorrunning{ECCV-20 submission ID \ECCVSubNumber} 
\author{Anonymous ECCV submission}
\institute{Paper ID \ECCVSubNumber}

\maketitle

In the supplementary material, we provide the PyTorch source code for implementing our proposed Attentive Normalization. The complete source codes for reproducing all the ImageNet experiments and COCO experiments are provided too. Please follow the readme in each code package. Due to space limit, we can not upload the pre-trained models, which will be made publicly available. 

We note that our codes are built on several existing code bases: the vanilla AOGNets~\cite{AOGNets}, different networks implemented in TorchVision, the MMDetection package~\cite{mmdetection}, and the NVIDIA APEX library. We will properly credit those code bases in our public release of the code.

\begin{lstlisting}[language=Python]
from __future__ import absolute_import
from __future__ import division
from __future__ import print_function
from __future__ import unicode_literals

import torch
import torch.nn as nn
import torch.nn.functional as F

_inplace = True
_norm_eps = 1e-5


### hsigmoid
class hsigmoid(nn.Module):
    def forward(self, x):
        out = F.relu6(x + 3, inplace=True) / 6
        return out


### Feature Norm
def FeatureNorm(norm_name, num_channels, num_groups, num_k, attention_mode):
    if norm_name == "BatchNorm2d":
        return nn.BatchNorm2d(num_channels, eps=_norm_eps)
    elif norm_name == "GroupNorm":
        assert num_groups > 1
        if num_channels % num_groups != 0:
            raise ValueError("channels {} not dividable by groups {}".format(
                              num_channels, num_groups))
        return nn.GroupNorm(num_channels, num_groups, eps=_norm_eps)
    elif norm_name == "AttentiveBatchNorm2d":
        assert num_k > 1
        return AttentiveBatchNorm2d(num_channels, num_k, attention_mode)
    elif norm_name == "AttentiveGroupNorm":
        assert num_groups > 1 and num_k > 1
        if num_channels % num_groups != 0:
            raise ValueError("channels {} not dividable by groups {}".format(
                              num_channels, num_groups))
        return AttentiveGroupNorm(num_channels, num_groups, num_k,
                                  attention_mode)
    else:
        raise NotImplementedError("Unknown feature norm name")


### Attention weights for attentive norm
class AttentionWeights(nn.Module):
    expansion = 2
    def __init__(self, attention_mode, num_channels, k,
                norm_name=None, norm_groups=0):
        super(AttentionWeights, self).__init__()
        self.k = k
        self.avgpool = nn.AdaptiveAvgPool2d(1)
        layers = []
        if attention_mode == 0:
            layers = [ nn.Conv2d(num_channels, k, 1),
                        nn.Sigmoid() ]
        elif attention_mode == 4:
            layers = [ nn.Conv2d(num_channels, k, 1),
                        hsigmoid() ]
        elif attention_mode == 1:
            layers = [ nn.Conv2d(num_channels, k*self.expansion, 1),
                        nn.ReLU(inplace=True),
                        nn.Conv2d(k*self.expansion, k, 1),
                        nn.Sigmoid() ]
        elif attention_mode == 2: # best, others for ablation study
            assert norm_name is not None
            layers = [ nn.Conv2d(num_channels, k, 1, bias=False),
                        FeatureNorm(norm_name, k, norm_groups, 0, 0),
                        hsigmoid() ]
        elif attention_mode == 5:
            assert norm_name is not None
            layers = [ nn.Conv2d(num_channels, k, 1, bias=False),
                        FeatureNorm(norm_name, k, norm_groups, 0, 0),
                        nn.Sigmoid() ]
        elif attention_mode == 6:
            assert norm_name is not None
            layers = [ nn.Conv2d(num_channels, k, 1, bias=False),
                        FeatureNorm(norm_name, k, norm_groups, 0, 0),
                        nn.Softmax(dim=1) ]
        elif attention_mode == 3:
            assert norm_name is not None
            layers = [ nn.Conv2d(num_channels, k*self.expansion, 1, bias=False),
                       FeatureNorm(norm_name, k*self.expansion, norm_groups, 0,
                                    0),
                       nn.ReLU(inplace=True),
                       nn.Conv2d(k*self.expansion, k, 1, bias=False),
                       FeatureNorm(norm_name, k, norm_groups, 0, 0),
                      hsigmoid() ]
        else:
            raise NotImplementedError("Unknow attention weight type")
        self.attention = nn.Sequential(*layers)

    def forward(self, x):
        b, c, _, _ = x.size()
        y = self.avgpool(x)
        var = torch.var(x, dim=(2, 3)).view(b, c, 1, 1)
        y *= (var + 1e-3).rsqrt() # RSD
        #y = torch.cat((y, var), dim=1) # for ablation study
        return self.attention(y).view(b, self.k)


### AN w/ BN
class AttentiveBatchNorm2d(nn.BatchNorm2d):
    def __init__(self, num_channels, k, attention_mode,
                 eps=_norm_eps, momentum=0.1, track_running_stats=True):
        super(AttentiveBatchNorm2d, self).__init__(num_channels, eps=eps,
            momentum=momentum, affine=False,
            track_running_stats=track_running_stats)
        self.k = k
        self.weight_ = nn.Parameter(torch.Tensor(k, num_channels))
        self.bias_ = nn.Parameter(torch.Tensor(k, num_channels))

        self.attention_weights = AttentionWeights(attention_mode, num_channels,
                                    k, norm_name='BatchNorm2d')

        self._init_params()

    def _init_params(self):
        nn.init.normal_(self.weight_, 1, 0.1)
        nn.init.normal_(self.bias_, 0, 0.1)

    def forward(self, x):
        output = super(AttentiveBatchNorm2d, self).forward(x)
        size = output.size()
        y = self.attention_weights(x) # bxk

        weight = y @ self.weight_ # bxc
        bias = y @ self.bias_ # bxc
        weight = weight.unsqueeze(-1).unsqueeze(-1).expand(size)
        bias = bias.unsqueeze(-1).unsqueeze(-1).expand(size)

        return weight * output + bias


# AN w/ GN
class AttentiveGroupNorm(nn.Module):
    __constants__ = ['num_groups', 'num_channels', 'k', 'eps', 'weight',
                     'bias']

    def __init__(self, num_channels, num_groups, k, attention_mode,
                 eps=_norm_eps):
        super(AttentiveGroupNorm, self).__init__()
        self.num_groups = num_groups
        self.num_channels = num_channels
        self.k = k
        self.eps = eps
        self.affine = True
        self.weight_ = nn.Parameter(torch.Tensor(k, num_channels))
        self.bias_ = nn.Parameter(torch.Tensor(k, num_channels))
        self.register_parameter('weight', None)
        self.register_parameter('bias', None)

        self.attention_weights = AttentionWeights(attention_mode, num_channels,
                                    k,
                                    norm_name='GroupNorm', norm_groups=1)

        self.reset_parameters()

    def reset_parameters(self):
        nn.init.normal_(self.weight_, 1, 0.1)
        nn.init.normal_(self.bias_, 0, 0.1)

    def forward(self, x):
        output = F.group_norm(
            x, self.num_groups, self.weight, self.bias, self.eps)
        size = output.size()

        y = self.attention_weights(x)

        weight = y @ self.weight_
        bias = y @ self.bias_

        weight = weight.unsqueeze(-1).unsqueeze(-1).expand(size)
        bias = bias.unsqueeze(-1).unsqueeze(-1).expand(size)

        return weight * output + bias

    def extra_repr(self):
        return '{num_groups}, {num_channels}, eps={eps}, ' \
            'affine={affine}'.format(**self.__dict__)




\end{lstlisting}

%
%
\bibliographystyle{splncs04}
\bibliography{egbib}